 \documentclass[10pt,twocolumn,letterpaper]{article}
 
\usepackage{cvpr}
\usepackage{times}
\usepackage{epsfig}
\usepackage{graphicx}
\usepackage{amsmath}
\usepackage{amssymb}
\usepackage{booktabs}
\usepackage{microtype}
\usepackage[numbered,framed]{matlab-prettifier}

\usepackage[pagebackref=true,breaklinks=true,letterpaper=true,colorlinks,bookmarks=false]{hyperref}

\cvprfinalcopy
\def\cvprPaperID{****}

\ifcvprfinal\fi

\begin{document}

\title{An Exploration of Neural Radiance Field Scene Reconstruction\\Synthetic, Real-world and Dynamic Scenes}

\author{
  Benedict Quartey
  \and
  Tuluhan Akbulut
    \and
  Wasiwasi Mgonzo
    \and
  Zheng Xin Yong
}

\maketitle

\begin{abstract}
This project presents an exploration into 3D scene reconstruction of synthetic and real-world scenes using Neural Radiance Field (NeRF) approaches. We primarily take advantage of the reduction in training and rendering time of neural graphic primitives multi-resolution hash encoding, to reconstruct static video game scenes and real-world scenes–comparing and observing reconstruction detail and limitations. Additionally, we explore dynamic scene reconstruction using Neural Radiance Fields for Dynamic Scenes(D-NeRF). Finally, we extend the implementation of D-NeRF, originally constrained to handle synthetic scenes to also handle real-world dynamic scenes.
\end{abstract}
\section{Introduction}


In recent years, rendering photo-realistic views of a scene become a popular research topic given the capability of neural rendering techniques. Neural radiance fields (NeRF) \cite{mildenhall2020NeRF} show a remarkable capacity to recover the 3D geometry and appearance from a sparse collection of 2D images of a scene taken from different angles. \cite{attal2021torf,pumarola2021dNeRF} even show that we can use NeRF-based algorithms to render scenes with dynamic objects. However, most work are done on scenes with synthetically generated objects created by softwares such as Blender. Therefore,  we are interested in the rendering capability of NeRF along two axis: (1) \textbf{static or dynamic} objects and (2) \textbf{synthetically-generated or real-life} objects.

Our project \footnote{The source code and videos of the results can found in appendix.} explores 3D scene reconstruction with Neural Radiance Field approaches. We are interested in 3D reconstruction from pre-captured scenes because of the unique ability to view already captured scenes from novel angles. We successfully use NeRF to model video game scene with synthetic character figures and real life scenes with objects and moving body parts. 
\\


\section{Problem Statement}
Traditional NeRF approaches can reconstruct both synthetic and real-world scenes and new methods like Instant Neural Graphics Primitives \cite{muller2022instant} significantly speed up the NeRF training process, however, these methods are limited to scenes with static Objects.
D-NeRF (Dynamic NeRF \cite{pumarola2021dNeRF}) extends traditional NeRF with time conditioning making it possible to reconstruct scenes with dynamic objects, however, the implementation of D-NeRF was limited to synthetic scenes where ground truth camera parameters exist. Our goal is to extend the implementation of D-NeRF to reconstruct real-world scenes with dynamic objects like dancing people.

\begin{figure*}[t!]
    \centering
    \includegraphics[width=\linewidth]{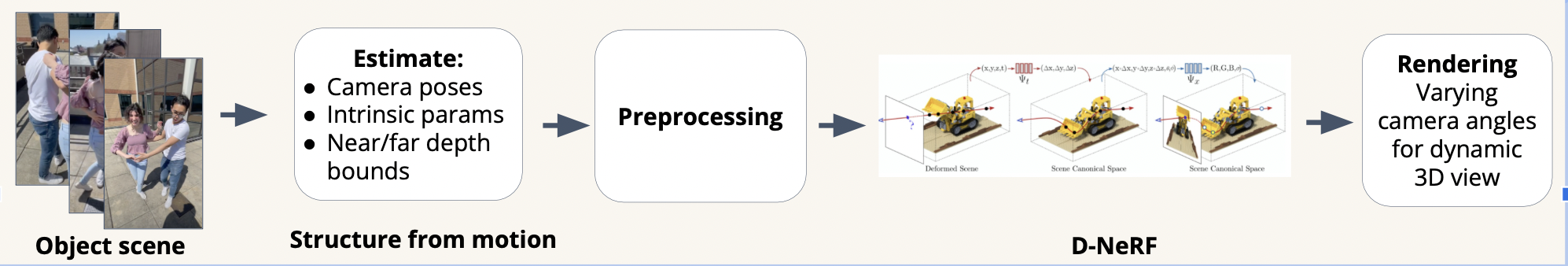}
    \caption{Proposed Architecture.}
    \label{fig:Proposed D-NeRF Architecture}
\end{figure*}

\section{Related Work}
Novel view synthesis is a long standing problem in computer vision. Recent approaches to solve the problem have used neural scene representations and rendering techniques that involves learning 3D scene representation from partial observations captured by sparse set of 2D images. One interesting application of this is to render novel views of a scene which finds real-world applications in augmented reality, 3D content production and animation of human bodies among many other applications. Despite these great promises, these representations are computationally expensive to train and often slow to render \cite{bergman2021fast}.\\
Recent approaches use coordinate-based neural representations combined with volume rendering. These approaches are called Neural Radiance Field (NeRF). NeRF are new approaches to rendering novel photo-realistic views of a scene from a sparse set of input images \cite{mildenhall2020NeRF,pumarola2021dNeRF}. This is done by using simple multilayer perceptron networks to encode mapping from spatial locations as well as camera views to emitted radiance values and volume density. This learned mapping from 5D inputs then allows free-viewpoint rendering with high-resolution geometry.\\
Although these approaches provide photo-realistic quality for synthesized image, they are only applicable to static scenes. Another shortfall is that they are slow to train and render\cite{mildenhall2020NeRF}. Other coordinate-based approaches model surface combined with sphere training based rendering, however these too suffer from the same problem that they are expensive to train \cite{riegler2021stable,oechsle2021unisurf}.
In addition, NeRF struggles with images obtained in uncontrolled settings. Therefore, \cite{martin2021NeRF} extend the work of NeRF by adding Generative Latent
Optimization and “transient” MLP head so that NeRF can model real-world phenomena in uncontrolled images including variable illumination or transient occluders. 

Further, because NeRF is also costly to train and evaluate due to its fully connected MLP layers, \cite{muller2022instant} creates Instant-NGP which uses a multiresolution hash table of trainable feature vectors to reduce the cost of NeRF rendering. This hash table permits using a smaller network without sacrificing rendering quality, so the number of floating point and memory access operations is lower. In our work, we use Instant-NGP for rendering of static scenes. 

For rendering dynamic scenes, D-NeRF extends NeRF idea to dynamic settings first by representing time-based scene deformation with a canonical configuration, and then by predicting the next volume density and RGB values using that configuration. Both mappings are simultaneously learned using fully-connected networks. Once the networks are trained, \texttt{D-NeRF} can render novel images, controlling both the camera view and the time variable, and thus, the object movement. However, this method requires synthetic scenes with ground truth camera parameters. In our work, we infer unknown camera parameters from real-world images and use D-NeRF to predict real-world dynamics.

\section{Method}
\subsection{Static Scene Rendering}
Given a set of images of the static scene, we used Instant-NGP to create the 3D representation of the scene in couple of minutes by exploiting the multiresolution hash table. The model requires parameters like aabb-scale, which determines extent of the scene, and transformation matrix for each image. These parameters are unknown for our video game video and real-world video. Therefore, we used \texttt{COLMAP} \cite{schoenberger2016sfm} to infer those parameters for sampled frames taken from videos. Then, the images and the found parameters are given to the model for rendering.

\subsection{Dynamic Scene Rendering}
Our main contribution is extending the implementation of D-NeRF to reconstruct real-world scenes with dynamic objects. To do this we developed a pipeline that takes a video of the scene as input, we extract frames from the video and estimate camera poses, intrinsic parameters and near/far depth bounds using structure from motion (we again take advantage of the \texttt{COLMAP} \cite{schoenberger2016sfm} package to do this). After this step, we preprocess this obtained information and compute timestamps for each camera pose to fit the structure accepted by the D-NeRF model. Additionally, we utilise preprocessing transformations from LLFF (Local Light Field Fusion)  as in \cite{mildenhall2020NeRF} to increase clarity. Then, we train D-NeRF to learn the 3D representation of the scene. 
The provided rendering module is creating a similar video to the original video with similar camera movements, which makes validation of 3D reconstruction harder. Therefore, we use rotation matrices along x and y axis to rotate the camera continuously to render novel views of the reconstructed scene. The whole architecture is in figure \ref{fig:Proposed D-NeRF Architecture}.\\


\section{Experimental Setting}
\subsection{Dataset Collection}
For static scenes with synthetically generated objects, we chose video game, particularly Defense of the Ancients 2. The reason is that the game enables us to zoom in and rotate to provide 360 views of the video game characters. For real-life object, we put a helmet on a chair and rotated the camera around it. For both scenes, we recorded a one-minute long video with 360 degree angles focusing on the subject of the scene, which was a video game character and the helmet on the chair.

For dynamic scenes, we used scenes with different intensity of body parts movement. The first scene was a human subject standing and moving his arms from waist height to shoulder height. In this scene, the only moving parts were both arms. On the other hand, the second scene had more body movements, involving two human subjects doing Lindy Hop dancing. In this scene, both subjects moved every single part of their body and away from the spot they originally stood at. We recorded one-minute videos for each scene while camera was rotated for 360 degrees around subjects.

\subsection{Hardware}
We used a single GPU (RTX 3090). For static settings, the training time was 5 minutes. For dynamic settings, it was 4 hours.

\section{Results}

\begin{figure}[t]
    \centering
    \includegraphics[width=\linewidth]{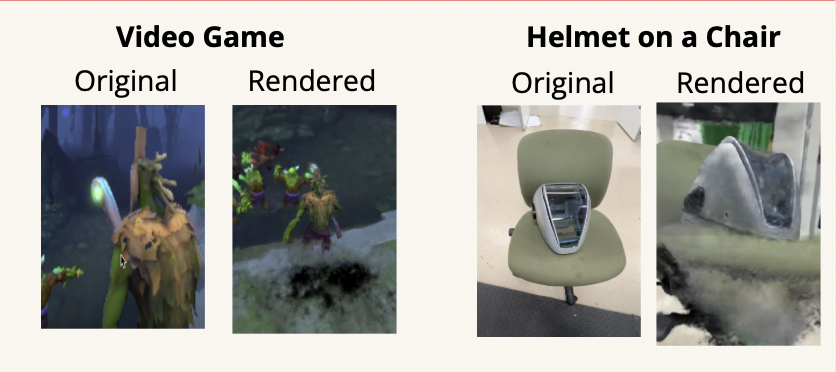}
    \caption{Static scene: Dota 2 and helmet on chair}
    \label{fig:result_static_scene}
\end{figure}
\begin{figure}[t]
    \centering
    \includegraphics[width=\linewidth]{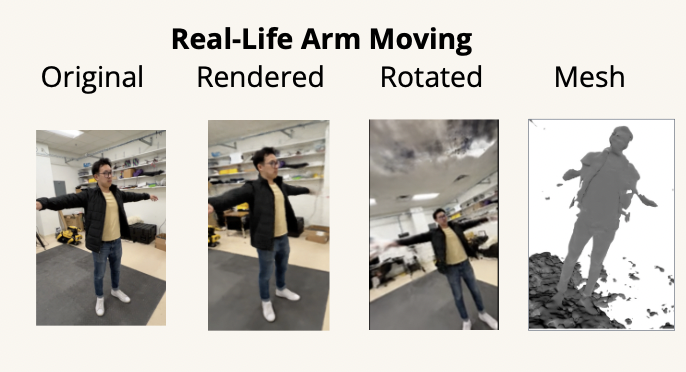}
    \caption{Dynamic real-life scene: Arm moving.}
    \label{fig:result_arm}
\end{figure}

\begin{figure}[t]
    \centering
    \includegraphics[width=\linewidth]{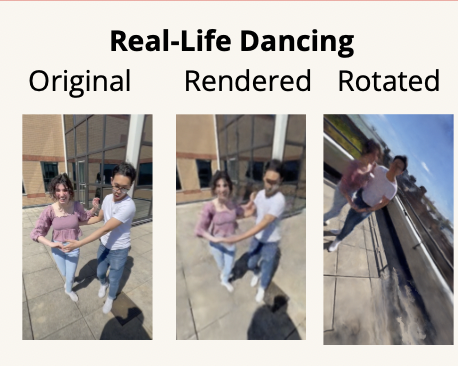}
    \caption{Dynamic real-life scene: Lindy Hop dancing.}
    \label{fig:result_dancing}
\end{figure}

\begin{figure}[t]
    \centering
    \includegraphics[width=\linewidth]{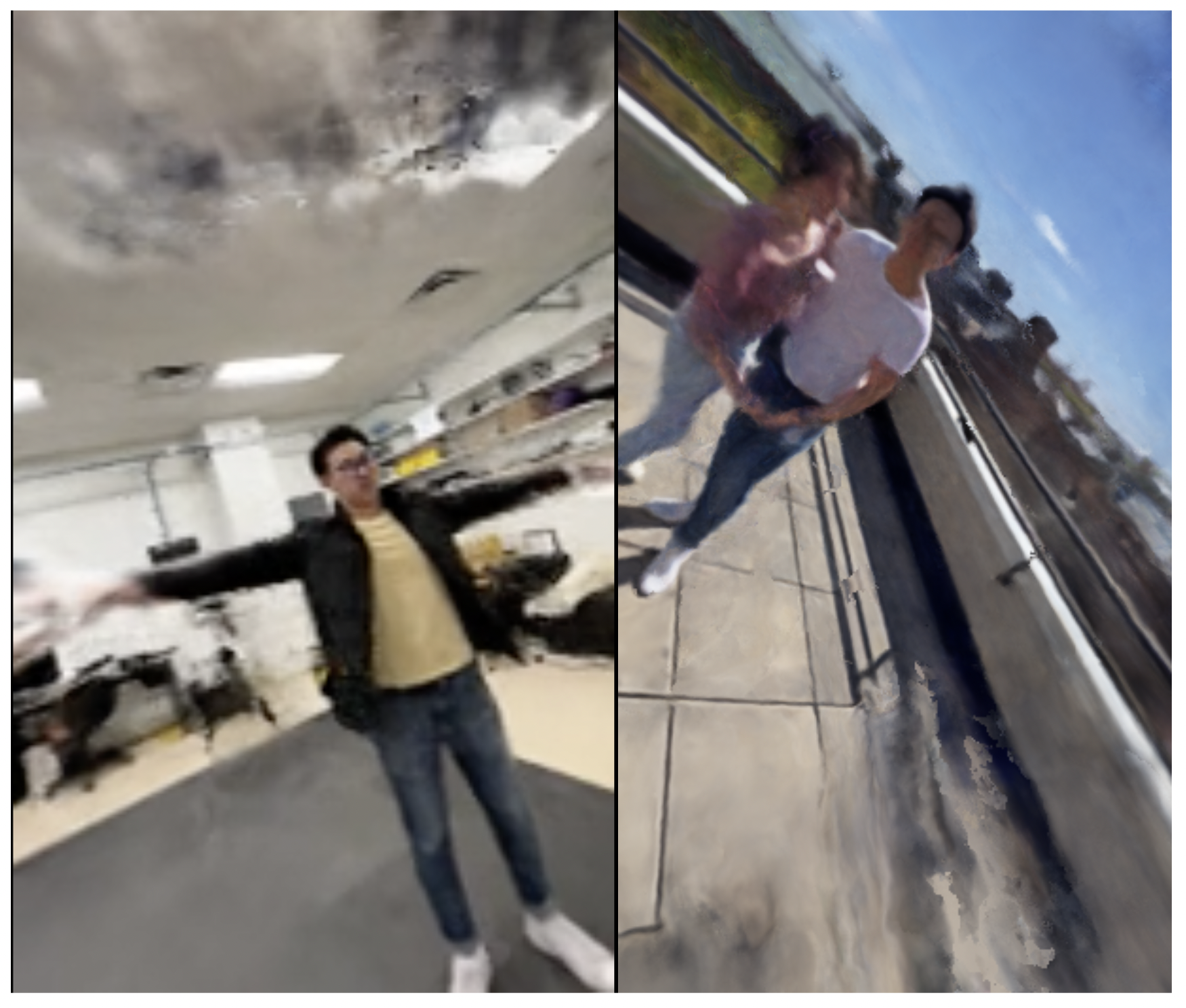}
    \caption{Comparison of the two dynamic video tests.}
    \label{fig:result_comparison}
\end{figure}
\begin{figure}[t]
    \centering
    \includegraphics[width=\linewidth]{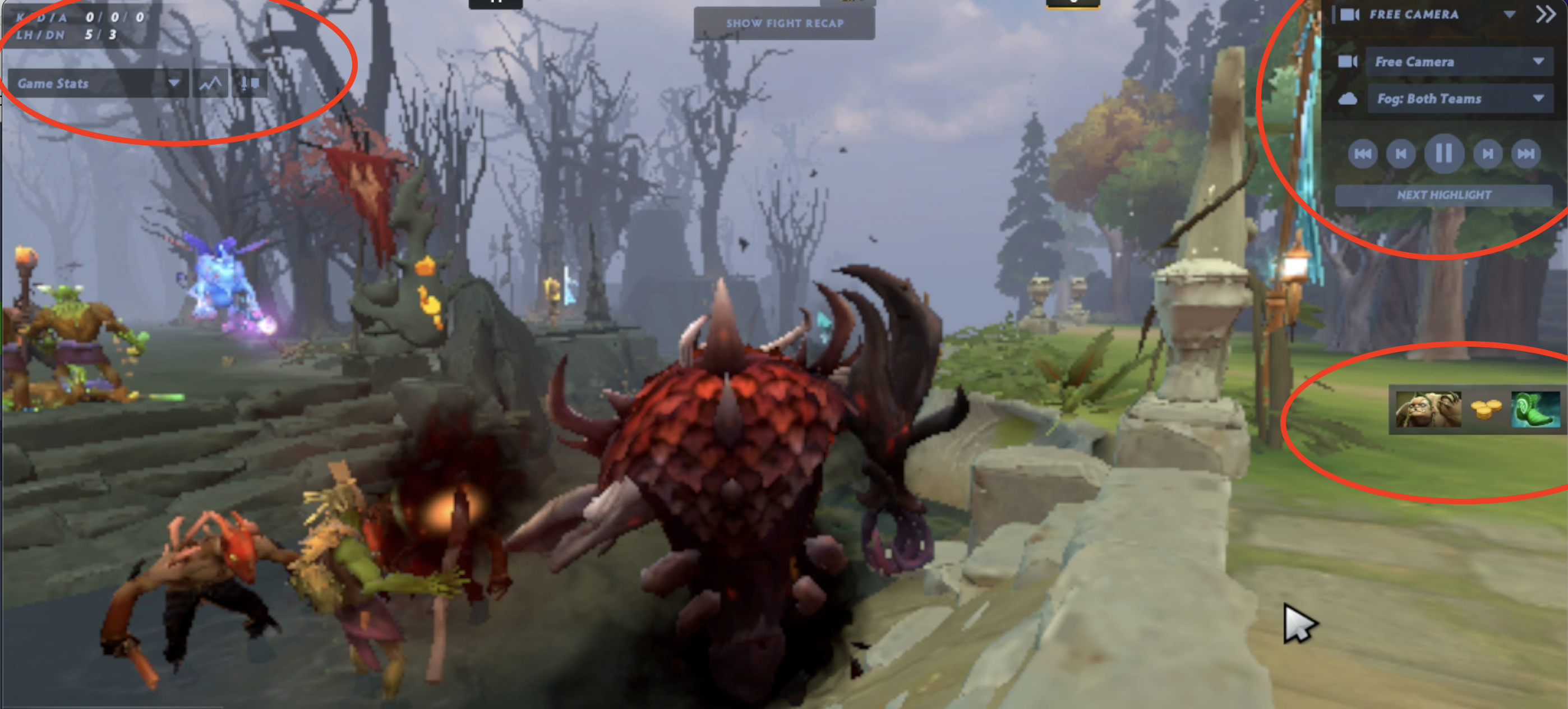}
    \caption{Dota 2 image data. Red circles indicate image components that lack depth.}
    \label{fig:result_dota2_circles}
\end{figure}
\begin{figure}[t]
    \centering
    \includegraphics[width=\linewidth]{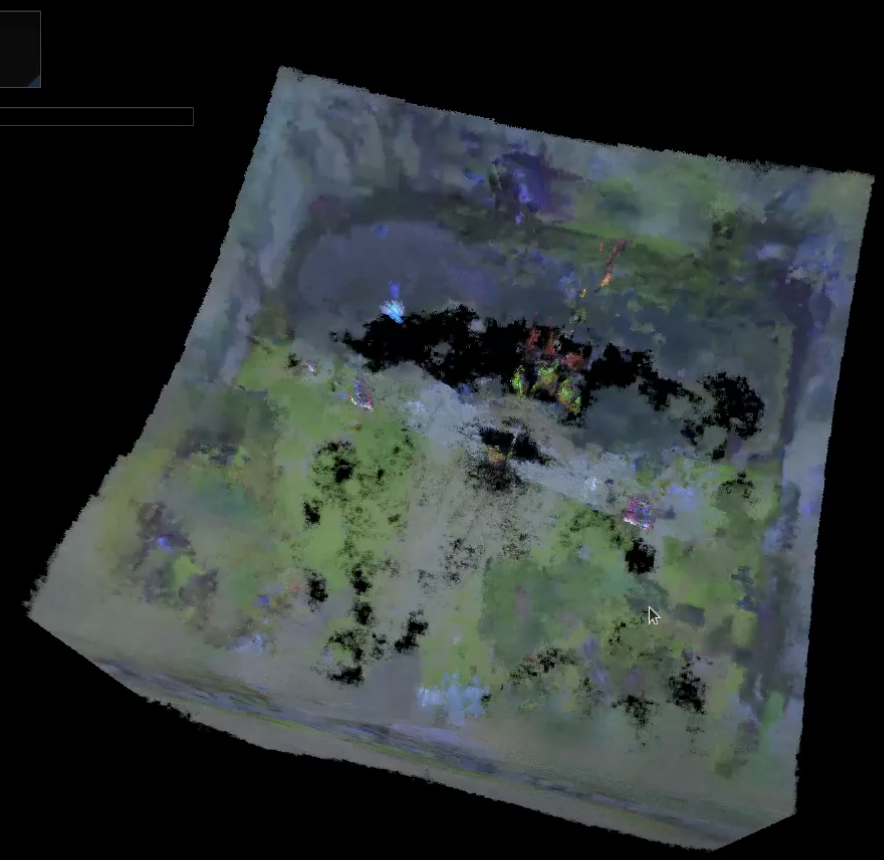}
    \caption{Rendered Dota 2 scene}
    \label{fig:result_dota2_rendered}
\end{figure}
\begin{figure}[t]
    \centering
    \includegraphics[width=\linewidth]{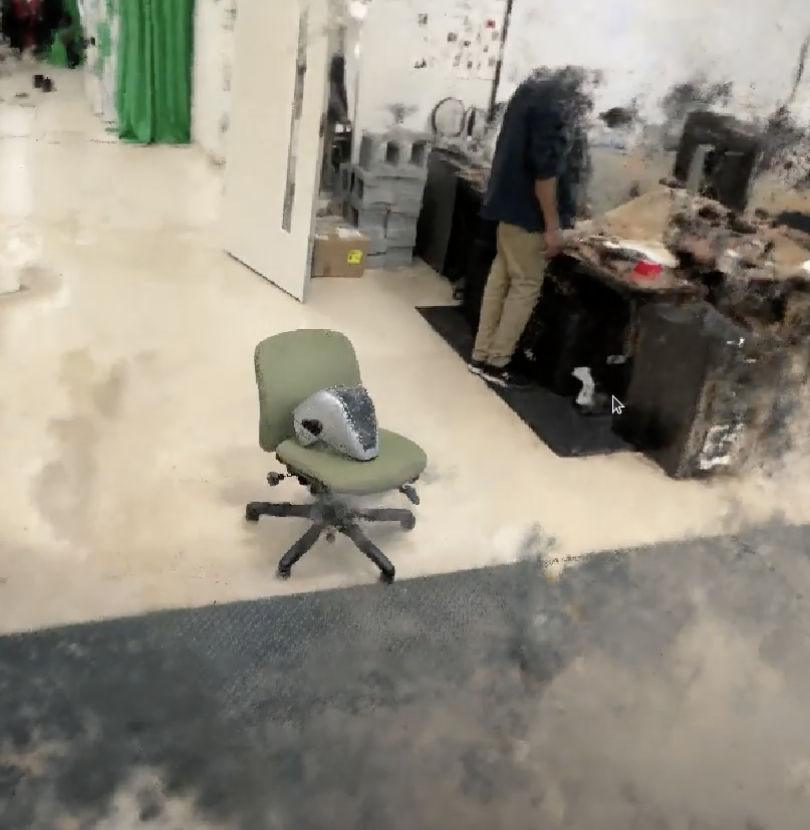}
    \caption{Rendered helmet on a chair scene.}
    \label{fig:result_helmet_rendered}
\end{figure}

\subsection{Technical Discussion}
\paragraph{Static scenes} For static scenes, we initially struggled with reconstructing video scenes. We encountered the error of "no pairing images found" during the interpolation of images for NeRF reconstruction. The main reason was that, as shown in Figure~\ref{fig:result_dota2_circles}, there are image components, such as tooltips that persists throughout the game scene, that lack depth. These floating panels in the images prevent interpolation of images for scene rendering, and our solution is to crop them out.

We successfully rendered the both video game scene and the real-life scene. Figure~\ref{fig:result_dota2_rendered} provides a panoramic view of the modeled video game scene, whereas Figure~\ref{fig:result_helmet_rendered} zooms in on the subject of the real-life scene. As observed in Figure~\ref{fig:result_helmet_rendered}, NeRF reconstructs not only the foreground subjects (i.e., the helmet and the chair) very clearly, but also able to models the background realistically such as the pant of a person standing and the shelves. The cloudiness in Figure~\ref{fig:result_helmet_rendered} is due to the low volume density at the ceiling level in the emitted radiance, as when we record the video of the helmet, our camera is at eye level. 

\paragraph{Dynamic scenes}


Figure \ref{fig:result_arm} illustrates the first dynamic video test where only the arms of the subject were moving. The model successfully rendered the video and the image with rotated camera validates the 3D reconstruction. Here, the mesh reconstruction does not represent time, therefore contains only static parts of the subject. 

Figure \ref{fig:result_dancing} shows the second dynamic video test where subjects were dancing. Here we also see that the model was able to render the videos. The image with rotated camera again illustrates the possibility of rendering with any desired camera angle. 

Figure \ref{fig:result_comparison} shows the differences in video qualities with respect to the amount of motion in videos. We observed that the faces of the subject was less clearer in the dancing video because they were moving whereas the face was static in the first video (arm movement). This tells that the network can not give similar accuracy for static and dynamic body parts, and the dynamics prediction needs to be improved for realistic video production.

As previously, the cloudiness in both reconstructions are due to the low volume density in the emitted radiance since videos are captured at eye level.

\subsection{Societal Discussion}

The first critique is about the misuse for modeling people without consent. Currently, the privacy laws such as Wiretap Act do not prevent people from recording others in public spaces. Hence, we should educate the users about the respect for privacy and consent. If we want to strictly prevent modeling humans in animations and video game avatars using recorded videos, we can incorporate human detection algorithms into the NeRF technology such that we can prevent rendering of human subjects in the scene. 

Another potential misuse is the creation of fake images. For instance, a political candidate can make a speech by superimposing themselves in an important institutional building (e.g. the White House) that was generated by NeRF. One way to prevent deepfake content creation is by adding a digital fingerprint or watermark to the input images so that it becomes more difficult for someone to create synthetic content from them. 

One concern raised is that military can use modeling technology to gain topographical information on a foreign area. However, we want to point out that the NeRF technology is quite limited in its ability as we cannot reconstruct the geographical and topographical scenes based on satellite images. It is also beyond our scope and knowledge as to how military is using neural reconstruction of geolocations as, similar to Google Maps reconstructing the street scene, NeRF enables reconstruction using phone cameras. In essence, NeRF behaves similarly to existing technology.

The fourth concern is the reduced funding for public spaces as we can now create 3D representations of major landmarks and locations so people may no longer visit the actual locations. We would like to push back on this point by suggesting the opposite: now with neural rendering of tourist destinations for marketing purposes, people have the opportunity to become aware of them so we can expect more tourists to visit the place in-person.

Privacy issue can be another concern. One critique is that NeRF technology can be abused to create unlicensed reproduction of 3D art, which breaks copyright and privacy laws of private company spaces and other important institutions. We believe that this is a very nuanced privacy issue because NeRF rendering requires 360 camera-filming on the artwork scene. In other words, if taking recorded videos or pictures of the real-life artwork doesn't violate the privacy issue, then by such the rendering shouldn't cause any societal harms. The direct way prevent infringement of copyright issues is to forbid any form of picture-taking or recording so reconstruction is not possible.

The last potential impact is the increase in the amount of spam on social media due to the misuse of NeRF to create massive amounts of distinct realistic pictures. We agree with the sentiment that users have another channel for sharing content. Previously, users can only share content in 2D images or videos, whereas now with NeRF, they have the ability to share content through neural rendering. While this may lead to a significant increase of content on social media, we believe that it also enables users to have more opportunities for capturing the memorable moments. For instance, when one moves to a new home, he or she can use neural rendering to share the interior designs of the house with friends and family.


\section{Conclusion}

In this project, we presented a pipeline on top of \texttt{D-NeRF} for modelling dynamic real-world scenes. While previous approaches have applied \texttt{D-NeRF} to synthetic dataset, we applied both to synthetic and real-world scene data. The results are interesting in that \texttt{D-NeRF} can render novel images, controlling both the camera view and the time variable, and thus, the object movement. Our approach opens new and promising research directions for 3D reconstruction. In the future, we can apply the technique to create novel 3D images where one can dynamically interact with the content at a particular point in time.

{\small
\bibliographystyle{plain}
\bibliography{ProjectFinal_ProjectReportTemplate}
}

\end{document}